# Towards proactive self-adaptive AI for non-stationary environments with dataset shifts

David Fernández Narro
*Biomedical Data Science Lab, Instituto Universitario de Tecnologías de la Información y Comunicaciones Universitat Politècnica de Valéncia*
Valencia, Spain
dfernar@upv.es

Pablo Ferri
*Biomedical Data Science Lab, Instituto Universitario de Tecnologías de la Información y Comunicaciones Universitat Politècnica de Valéncia*
Valencia, Spain
pabferb2@upv.es

Juan M García-Gómez
*Biomedical Data Science Lab, Instituto Universitario de Tecnologías de la Información y Comunicaciones Universitat Politècnica de Valéncia*
Valencia, Spain
juanmig@upv.es

Carlos Sáez
*Biomedical Data Science Lab, Instituto Universitario de Tecnologías de la Información y Comunicaciones Universitat Politècnica de Valéncia*
Valencia, Spain
carsaesi@upv.es

*Abstract*— Artificial Intelligence (AI) models deployed in production frequently face challenges in maintaining their performance in non-stationary environments. This issue is particularly noticeable in medical settings, where temporal dataset shifts often occur. These shifts arise when the distributions of training data differ from those of the data encountered during deployment over time. Further, new labeled data to continuously retrain AI is not typically available in a timely manner due to data access limitations. To address these challenges, we propose a proactive self-adaptive AI approach, or pro-adaptive, where we model the temporal trajectory of AI parameters, allowing us to short-term forecast parameter values. To this end, we use polynomial spline bases, within an extensible Functional Data Analysis framework. We validate our methodology with a logistic regression model addressing prior probability shift, covariate shift, and concept shift. This validation is conducted on both a controlled simulated dataset and a publicly available real-world COVID-19 dataset from Mexico, with various shifts occurring between 2020 and 2024. Our results indicate that this approach enhances the performance of AI against shifts compared to baseline stable models trained at different time distances from the present, without requiring updated training data. This work lays the foundation for pro-adaptive AI research against dynamic, non-stationary environments, being compatible with data protection, in resilient AI production environments for health.

## I. INTRODUCTION

Recent advances in Artificial Intelligence (AI) technologies and regulations make more accessible production-ready data-driven decision support based on learning from large volumes of complex data. However, when used in non-stationary environments, such as in health, AI models face a persistent challenge: the inherent temporal variability of real-world data, where data distributions at model use can differ from those observed at training, leading to erroneous decisions and predictions [1] due to potential model obsolescence. These phenomena are known as dataset shifts [2], which are typically disaggregated into prior probability shift, covariate shift, and concept shift.

In the medical domain, dataset shifts can manifest for various reasons, such as the introduction of new technologies, the emergence of novel diseases, changes in treatment protocols, or changing patient population composition [1]. These fluctuations in data are especially problematic in clinical settings, where they can severely impact the accuracy and reliability of AI models, leading to diagnostic or therapeutic errors that compromise patient safety [3], [4], [5].

Various approaches have attempted to mitigate the effects of dataset shifts through self-adaptive approaches such as continual learning [6] or domain adaptation techniques [7]. These solutions can be expensive, both in terms of computational and economic resources, and they are not always accessible in real time. This is particularly evident in clinical settings, where quick and accurate decision-making is crucial. Additionally, AI adaptation requires new labeled data and appropriate access permissions, which can complicate the process further.

In response to these limitations, we suggest that the variability of the machine learning model's covariates, outcomes, and parameters could be modeled through the kinematics of their trajectories over time, enabling forecasting of changes based on time series and Functional Data Analysis. This methodology has the potential to uncover the underlying dynamics in both data and models, enabling the estimation of future changes in these parameters even in the absence of new training data. In this way, we suggest it is possible to anticipate the response to gradual dataset shifts and accelerate the adaptation to abrupt shifts.

The proposed pro-adaptive AI approach aims to enhance the performance, resilience, and endurance of AI systems in medicine through well-informed and timely clinical decisions when used in real-world non-stationary environments, where decisions based on outdated models can significantly impact patient health.

## II. METHODS

### A. Data Collection and Preprocessing

For this study, we used two complementary datasets to evaluate our methodology. First, we generated a simulated dataset $\{x,y\}^N$ designed to capture controlled temporal variability. This dataset consists of a binary label y and a numerical variable x generated from two Gaussian distributions, each representing a distinct class y₁ and y₂. The means of these distributions alternate gradually over a four-year period, resulting in linear separability at the beginning and end of the time span, but with significant overlap in the middle. In consequence, the dataset contains a gradual covariate shift in p(y), a covariate change in p(x), and a significant and alternating concept shift, changing p(x|y). This controlled simulation allows us to assess our approach's capability to detect and adapt to shifts in dataset characteristics.

Secondly, we used the real-world COVID-19 dataset collected by the General Epidemiology Directorate of the Mexican Ministry of Health, comprising daily updated data from suspected COVID-19 cases in public and private hospitals nationwide since 2020. The data is anonymized, open-access, and published by the Mexican government [8]. The use of data followed Mexico's terms of free use of the Open Data of the Mexican Government. We combined the existing labels to classify SARS-CoV-2 cases as confirmed positive and non-confirmed cases. For this study, we selected a random sample of 1.000.000 cases spanning from 2020 to 2024. Although the original dataset included 42 variables, we retained 20 categorical variables after an exploratory data analysis chosen for their relevance in predicting COVID-19 positive cases using Logistic Regression models [9], along with the admission date. As data quality control, we removed duplicate rows, incomplete records such as those with NaN values in the label, and date inconsistencies.

In both datasets, we batched cases into yearly quarters based on dates. To ensure statistical robustness in the study, we performed a stratified by-class bootstrapping with 100 bootstrap samples for each temporal batch. This approach balanced statistical representativeness and computational cost, ensuring that data from all classes and temporal batches were well represented. For each temporal batch, we split the data into training and test sets following an 80-20 partition, while we further partitioned the training set into pure training and internal validation for hyperparameter tuning following a 70-30 split. Finally, we applied bootstrap sampling to the pure training set. In the COVID-19 dataset, we encoded the categorical variables in each subset using feature hashing, which maps them into a consistent 100-dimensional space. This dimensionality effectively captures the necessary information of the data, avoiding harmful collisions while keeping the feature space manageable [10]. This approach also addresses missing values and maintains consistent dimensionality across different temporal subsets, regardless of variability in category values.

*B. Temporal Variability Characterization*

Before AI modeling, to delineate the shifts occurring in the datasets, we performed an unsupervised characterization of the temporal variability of the data. We used the Python dashi tool [11] to calculate the data's Information Geometric Temporal (IGT) [12], which analyses the similarity between the data temporal batches' probability density function.

For the simulated dataset, to delineate the changepoints for a comparative basis while providing a clear view of shifts in a controlled environment, we analyzed the marginal probability distribution p(x) and the conditional distribution p(x|y) throughout the four-year period. Then, we computed the IGT for both p(x) and p(x|y), enabling us to study the dynamic changes in the feature space and its dependency on class labels.

For the COVID-19 dataset, we initially investigated the prior probability shift by examining the temporal distribution of COVID-19 prevalence p(y) and its IGT. To characterize covariate and concept shifts, we conducted a Multiple Correspondence Analysis (MCA) [13] over the categorical variables for multivariate analysis. In the case of covariate shift, we conducted MCA on the whole dataset and then applied a Kernel Density Estimation (KDE) on the first three principal components, which together accounted for 16.28% of the variance, sufficient to our purpose to perform a 3D exploratory analysis, allowing us to effectively visualize and explore the principal underlying shifting patterns. We used this information to estimate the joint probability density function and calculate the IGT. We performed MCA, retaining the three principal components for the concept shift again, and separated the data based on the COVID-19 positive cases label. Then, we applied the KDE to each subset to obtain the joint conditional probability density functions. After that, we concatenated the probability density functions of each class to calculate the IGT and evaluate the effect of the class on temporal variability.

*C. Baseline Predictive Models Development*

For the classification task, we selected logistic regression models due to their inherent interpretability [14], which facilitates the independent extraction and analysis of model parameters over time. Using Python PyTorch, we trained each model incrementally using training sets from consecutive temporal batches for each bootstrap sample. This state-of-the-art methodology enables models to adapt to incoming data in real time without complete retraining, updating model parameters incrementally as new data are received [15]. We employed a Bayesian hyperparameter optimization to systematically identify each model's optimal set of hyperparameters using the Optuna software [16]. For these hyperparameters tunning, we used the validation set corresponding to the last temporal batch used in training, with the primary performance metric being the Area Under the Receiver Operating Characteristic Curve (ROC-AUC). After selecting the best hyperparameters, we tested the models on test data encompassing all time periods to analyze performance stability and adaptability over time. Performance was assessed using ROC-AUC, recall, and macro-averaged F1-Score, ensuring that the chosen training strategy performed well on recent data and maintained robust performance across varying timestamps.

*D. Development of Pro-Adaptive AI Models with Times Series Forecasting*

After training each bootstrap model over time, we retained the temporal evolution of their parameters, treating them as independent time series. We used these time series to to effectively model the trajectories of parameters over time. To this end, we employed Functional Data Analysis (FDA), treating each time series as a distinct functional entity [17]. FDA provides a robust framework for analyzing data that vary continuously over a domain, such as time, by representing them as smooth functions rather than discrete points. This approach facilitates the extraction of underlying patterns and trends, enhancing both interpretability and predictive capability.

In this work, within the FDA framework, we employed polynomial splines as basis functions to capture the dynamics of each parameter's trajectory. We chose polynomial splines for their inherent flexibility, providing a balance between their capacity to model complex, nonlinear relationships within the data [17] and training simplicity. To determine the optimal degree of the spline, we conducted a cross-validation process across successive time periods, ensuring that the model adequately captures the data's variability without overfitting. Predictions for the subsequent temporal batches were generated based on the fitted splines, and their accuracy was assessed using the Mean Absolute Percentage Error (MAPE) for each parameter. This metric clearly measures the model's predictive performance over time, enabling us to evaluate and refine the spline configurations iteratively.

### E. Comparative of Baseline vs Pro-Adaptive Models

Once all the pro-adaptive models' parameters were forecasted, we evaluated their effectiveness by comparing the performance with the previously trained baseline models over three scenarios, comparing the mean and 95% confidence intervals (CI) over the bootstrap samples. In the first scenario, we used the baseline models trained on data up to the time t - Δ (last date with available data) and tested at time t (present). This simulates a real-world deployment environment where retraining or newly labeled data are not continuously available. In the second scenario, which corresponds to our proposed methodology, we used the pro-adaptive models with training data up to t - Δ, and we forecasted the parameters and tested their performance at time t. Finally, we compared a third scenario where the baseline models were trained and tested on the data at time t, which provided an upper-bound performance benchmark.

## III. RESULTS

### A. Temporal Variability Characterization

Figure 1 and Figure 2 show the temporal variability characterization of both the simulated and COVID-19 datasets, respectively.

In Figure 1, the left panel shows the evolution of the marginal probability distribution p(x), confirming that both Gaussian distributions suffer a gradual shift over time. The middle panel depicts the conditional probability distribution p(x|y), showing the shifts alternating the two Gaussian distributions mean generating each class, indicating that examining the conditioned distribution can be crucial for identifying concept shifts. The right panel presents the IGT plot of the conditioned probability distribution, confirming both gradual and abrupt shifts in the simulated data. These observations demonstrate that while overall changes may appear smooth, specific interactions between features and labels can introduce sudden transitions.

In Figure 2, the left panel shows the distribution of the COVID-19 dataset label p(y), positive cases, over time, reflecting the characterization of prior probability shifts over time, highlighting a dramatic fall in the probability of the negative class in October 2023. This may indicate the presence of a prior probability shift. The middle and right panels display the IGT of the covariates themselves and conditioned by label over time batched by month respectively, as an identifier of covariate and concept shift. Both situations present similar results, with 3 well-identified temporal subgroups. First, we identify some gradual changes over the 2020 data, followed by an abrupt change that results in a different temporal subgroup with data from the beginning of 2021 until the end of 2023. Finally, both IGT projections show another abrupt change, resulting in another temporal subgroup composed of data from 2024. These results indicate the presence of both covariate and concept shifts. These findings collectively emphasize the complex temporal dynamics within our data and highlight the need to consider abrupt and gradual shifts when analyzing temporal variability.

### B. Comparison of trained vs pro-adaptive estimated models

In Figures 3 and 4, we compare the performance of the baseline and pro-adaptive models across the three scenarios outlined in the methods section for the simulated and COVID-19 datasets, respectively. The green series relates to the upper-bound benchmark scenario. Blue series relates to the real-world baseline model, where we trained with data up to t – Δ and tested at time t. In both Figures 3 and 4, for the upper row, Δ is set to 2-temporal experiences (equivalent to six months), and for the lower row, Δ is 4-temporal experiences (equivalent to one year). The red series relates to the pro-adaptive models' performance with the same Δ as the blue series.

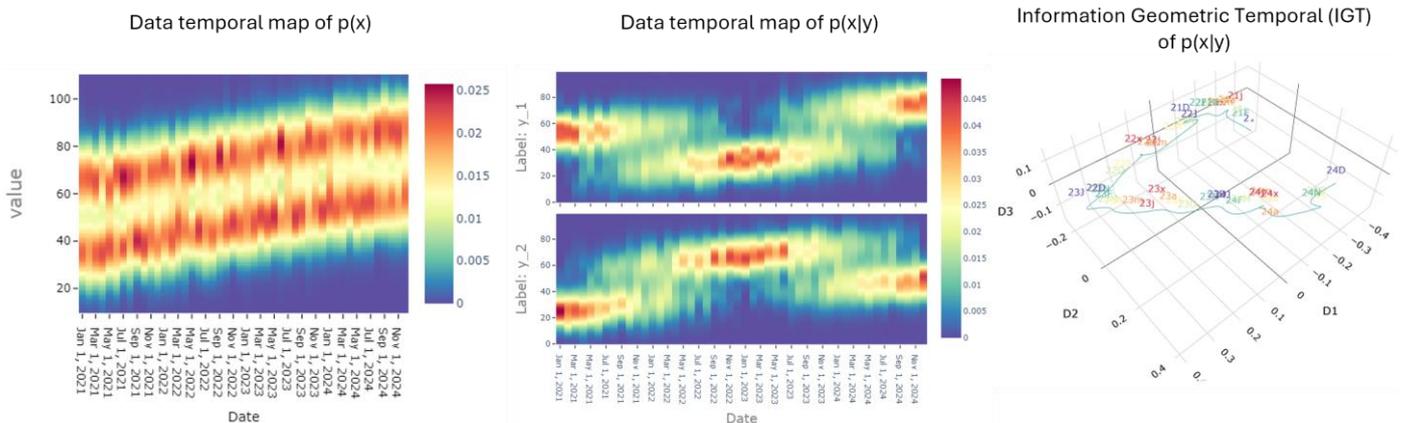

Fig. 1. Temporal Variability Characterization using the Python dashi tool for the Simulated Dataset. The left panel shows the evolution of the marginal probability distribution p(x) over time, with the x-axis representing timestamps and the y-axis the probability values. The middle panel displays the evolution of the conditional probability distribution p(x|y) separately for each class. The right panel presents the Information Geometric Temporal (IGT) representation of the conditioned probability distributions, highlighting both gradual shifts when the Gaussian distributions alternate.

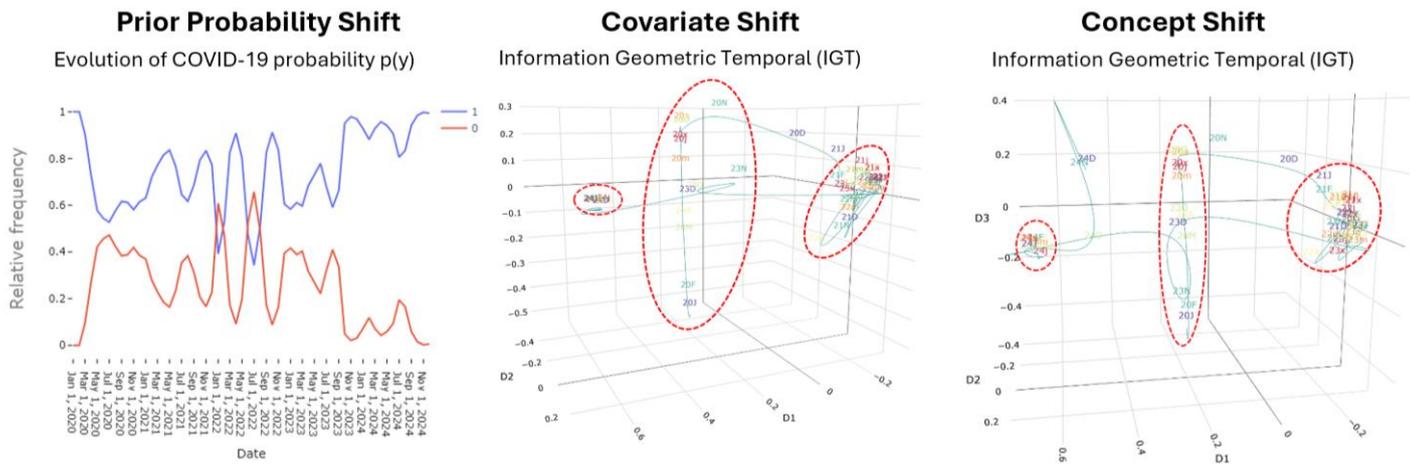

Fig. 2. Temporal Variability Characterization using the Python dashi tool for the COVID-19 dataset. The left panel shows the evolution of the probability distribution p(y) for COVID-19 positive cases over time. The middle panel presents the Information Geometric Temporal (IGT) of the covariates' probability distribution p(x) after dimensionality reduction via Multiple Correspondence Analysis (MCA). This panel reveals three distinct temporal subgroups, in red ovals, indicating the presence of both abrupt and gradual shifts, and highlighting a covariate shift in the data. The right panel displays the IGT of the conditioned probability distribution p(x|y), where the red ovals delineate the three temporal subgroups found associated to a concept shift.

The upper-bound benchmark (green series) is characterized in both datasets by a consistently high macro F1-Score, Recall, and ROC-AUC, showing a robust handling of dataset shifts. The simulated dataset presents smoother transitions associated with the alternate of the Gaussian distributions and the dataset shifts described in Figure 1. On the COVID-19 dataset, F1-Score and Recall metrics show a big CI with a decrease in the metrics in the last quarter of 2023. This is likely linked to the prior probability shift detected at the end of 2023 in Figure 2.

Regarding the baseline models (blue series), in both datasets, we observe that the model performance declines noticeably during the periods of dataset shifts characterized above. For the simulated dataset, this occurs when the two Gaussian distributions overlap more heavily. In the COVID-19 dataset, this decrease particularly happens in the first quarter of 2021 and the first two quarters of 2024.

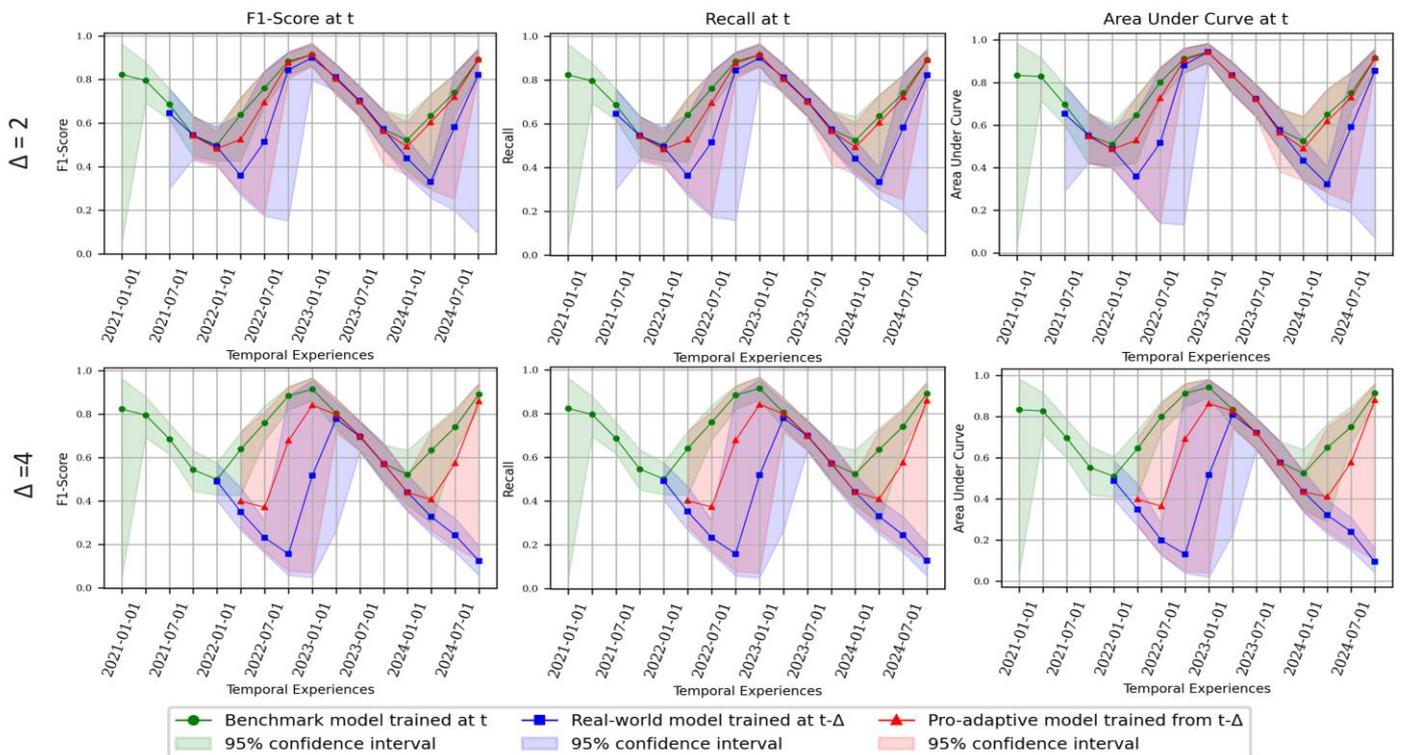

Fig. 3. Performance Metrics Comparison for the Simulated dataset. The green series represents models trained and tested with the most up-to-date data (ideal scenario). The blue series shows performance in a realistic scenario where training data is available only up to t − Δ; the upper row corresponds to Δ=2 temporal experiences (six months) and the lower row to = 4 temporal experiences (one year). The red series shows the performance of the pro-adaptive models trained using Functional Data Analysis with polynomic splines basis based on training data available up to t − Δ.

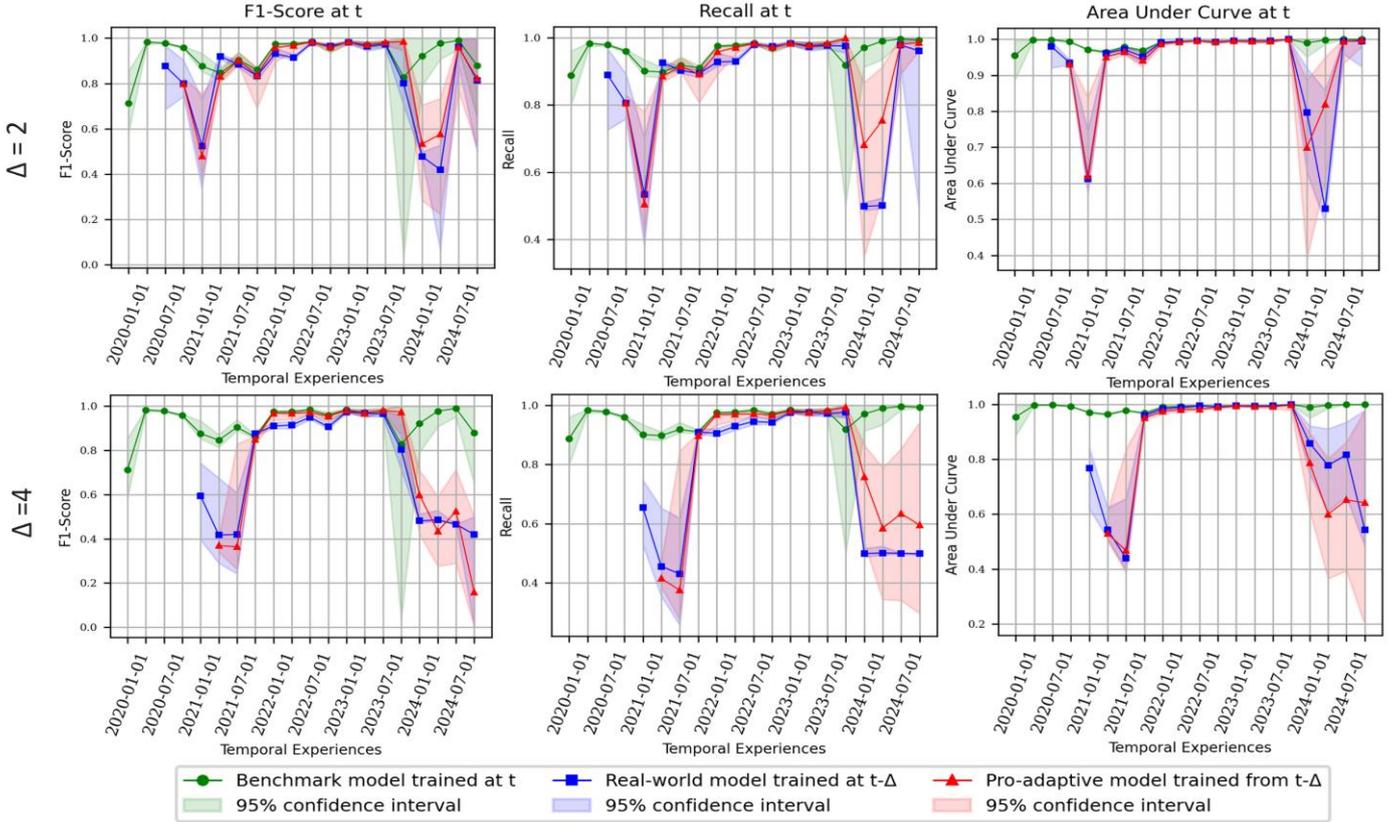

Fig. 4. Performance Metrics Comparison for the COVID-19 dataset. The green series represents models trained and tested with the most up-to-date data (ideal scenario). The blue series shows performance in a realistic scenario where training data is available only up to $t − \Delta$; the upper row corresponds to $\Delta=2$ temporal experiences (six months) and the lower row to $\Delta = 4$ temporal experiences (one year). The red series shows the performance of the pro-adaptive models trained using Functional Data Analysis with polynomic splines basis based on training data available up to $t − \Delta$.

Regarding the pro-adaptive models, in both datasets, they tend to closely track the performance of the upper-bound benchmark models, proactively self-adapting to temporal trends with sufficient accuracy in advance and with no use of newly labeled data. In the simulated dataset, the pro-adaptive model demonstrates a faster recovery from gradual shifts that coincide with the points of maximum overlap. This behavior is observed during the last quarter of 2021 and the first quarter of 2024, whereas the real-world environment model does not recover as quickly. In the COVID-19 dataset, we observed a slight overperformance at the abrupt shift of 2024 and a faster performance recovery for $\Delta = 2$ compared to the real-world environment model.

## IV. DISCUSSION

Our study's results show that our pro-adaptive AI estimation approach can effectively track changes and adapt AI models to non-stationary working environments. This approach applies forecasted model parameters through FDA using polynomial splines as a potential alternative to continuous retraining, particularly in situations where new data may not be immediately available. We observe that models trained with the most up-to-date data deliver optimal performance, as expected. However, in a realistic setting where retraining is delayed, our method maintains robust performance, capturing temporal trends and mitigating the impact of dataset shifts. This finding is particularly significant when working with real-world datasets, which often present shifts due to changes in clinical practices, population conditions, or health information systems. In the COVID-19 dataset, temporal variability may be attributed to evolving clinical classification systems and disease variants. This variability results in not only a prior probability shift, with gradual changes throughout the temporal period but also covariate and concept shifts, as evidenced by the identification of three temporal subgroups with gradual internal changes. These shifts translate into a noticeable decrease in AI model performance in the first quarter of 2021, when the pro-adaptive model lacks sufficient historical data to adjust. Conversely, during the last quarter of 2023 and the first quarter of 2024, the pro-adaptive model demonstrates generally superior behavior compared to the real-world deployment model.

Beyond performance metrics, our study offers insights into the potential benefits of integrating temporal forecasting techniques into AI systems to address dataset shifts proactively. By treating model parameters as time series and analyzing their trends over time, our work offers a framework for creating proactive self-adaptive AI systems. These systems can proactively adjust to changes in the data landscape, enhancing their resilience in non-stationary environments [18], [19]. This is especially valuable in healthcare, where re-training data availability may be intermittent due to delayed ground-truth assignments or lack of accession permits, and delayed adaptations can lead to diagnostic errors or suboptimal treatment recommendations [1]. In contrast to existing solutions that rely heavily on continuous retraining, our approach offers a resilient and cost-effective alternative. The comparative analyses on both a controlled simulated dataset and a real-world COVID-19 dataset underscore this versatility, demonstrating the method's ability to handle both gradual and abrupt changes.

It is worth mentioning that current European AI regulations limit and subject the modifications of self-adaptive AI to new conformity assessments or based on well-planned retraining criteria [1], [20], [21]. The fact of proactively adapting AI, as we propose, might be even more challenging from a regulatory perspective, particularly in high-risk medical systems. But considering the potential benefit for patients, further work should study this technology from the regulatory perspective, e.g., by using pro-adaptive AI models as supporting evidence complementary to conformant production AI models suggesting the review of uncertain cases.

Despite these promising results, several challenges and opportunities for future research remain. First, our study focuses primarily on logistic regression models, chosen for their interpretability and relatively manageable parameter space. Adapting this forecasting-based approach to models with numerous parameters, such as deep neural networks, could significantly expand its applicability, but would also require careful consideration of computational costs and the complexity of parameter interactions. Second, while we demonstrated the method's effectiveness on two datasets, additional experiments on more diverse datasets and domains would further validate its generalizability. Finally, exploring complementary forecasting techniques, such as other FDA families, deep learning time series forecasting techniques, or state-space models, could refine the accuracy and robustness of parameter prediction.

## V. CONCLUSIONS

We present a novel methodology that leverages Functional Data Analysis with polynomial splines to forecast AI model parameters, enabling our models to adapt to the challenges posed by real-world, non-stationary environments.

Our experiments on both the simulated and the COVID-19 datasets confirm that our pro-adaptive AI approach maintains robust performance under delayed data conditions when compared to real-world deployment baseline models, closely tracking upper-bound benchmarks by tracking and adjusting to the temporal dynamics inherent in the data. By treating model parameters as time series, our framework captures gradual shifts and enables an informed recovery from abrupt shifts. It also provides a resilient and cost-effective alternative to continuous retraining, which is particularly valuable in health settings where real-world data is intrinsically non-stationary and subject to changes, and AI re-training can be limited from delayed access to updated labeled data.

ACKNOWLEDGMENTS

This work was funded (PID2022-138636OA-I00; KINEMAI) by Agencia Estatal de Investigación—Proyectos de Generación de Conocimiento 2022.